\documentclass[11pt]{article}

\PassOptionsToPackage{table}{xcolor}
\usepackage[final]{acl}

\usepackage{times}
\usepackage{latexsym}

\usepackage[T1]{fontenc}
\usepackage[utf8]{inputenc}

\usepackage{microtype}

\usepackage{inconsolata}

\usepackage{graphicx}
\usepackage[most]{tcolorbox}
\usepackage{setspace}
\usepackage{booktabs}
\usepackage{makecell}
\usepackage{fancyvrb}
\usepackage{amsmath}
\usepackage{bm}

\newtcolorbox{promptbox}{
    breakable,
    colback=gray!10, colframe=black, arc=5mm, boxrule=0.5pt
}

\VerbatimFootnotes

\definecolor{correct}{rgb}{0.7, 0.9, 0.7}  % Lighter Green for correct answers
\definecolor{incorrect}{rgb}{1.0, 0.7, 0.6} % Lighter Red for incorrect answers

\def\vr{{\bm{r}}}
\def\vx{{\bm{x}}}
\def\vq{{\bm{q}}}

\def\va{{\bm{a}}}
\def\vs{{\bm{s}}}
\def\gV{{\mathcal{V}}}

\title{Reasoning Strategies in Large Language Models: Can They Follow, Prefer, and Optimize?}

\author{
  Yanjian Zhang$^{1,2}$ \quad
  Guillaume Wisniewski$^{2}$ \quad
  Nadi Tomeh$^{1}$ \quad
  Thierry Charnois$^{1}$ \\
  $^1$ Université Sorbonne Paris Nord, CNRS, LIPN, F-93430 Villetaneuse, France \\
  $^2$ Université Paris Cité, LLF, CNRS, 75013 Paris, France \\
  \texttt{\{yanjian.zhang,nadi.tomeh,thierry.charnois\}@lipn.univ-paris13.fr} \\
  \texttt{guillaume.wisniewski@u-paris.fr}
}

\begin{document}
\maketitle
\begin{abstract}
Human reasoning involves different strategies, each suited to specific problems. Prior work shows that large language model (LLMs) tend to favor a single reasoning strategy, potentially limiting their effectiveness in diverse reasoning challenges. In this work, we investigate whether prompting can control LLMs reasoning strategies and assess its impact on logical problem-solving. While our experiments show that no single strategy consistently improves accuracy, performance could be enhanced if models could adaptively choose the optimal strategy. We propose methods to guide LLMs in strategy selection, highlighting new ways to refine their reasoning abilities.
\end{abstract}

\section{Introduction}

Large-language models (LLMs) have exhibited impressive reasoning abilities when prompted with carefully designed instructions. Chain-of-thought (CoT) prompting, for instance, can elicit step-by-step deductions that markedly improve performance on mathematical, commonsense, and symbolic tasks \cite{Wei2022ChainOT,Ling2017ProgramIB, Cobbe2021TrainingVT,Kojima_Gu_Reid_Matsuo_Iwasawa_2022}. Yet most prompting techniques apply a single reasoning style---typically some variant of CoT---to every instance, whereas human problem-solvers dynamically choose among multiple strategies.

Cognitive science shows that people can switch, for example, between supposition following (hypothesising an assumption and tracing its consequences) and chain construction (building a sequential argument), selecting whichever suits the problem at hand \cite{vanderhenst02strategies, newton2004methods, johnson2010mental, Khemlani2019whymachine, Eisape2023ASC, Opedal2024DoLM,  Choi2024PeopleWA, NEURIPS2024_01025a4e, bao-etal-2025-likely}.
Moreover, some strategies are better suited to certain types of problems, and individuals may develop preferences for specific strategies based on their experience or cognitive abilities. In many cases, an expert's skill is precisely their ability to select the most appropriate strategy for solving a given problem.

This contrast between humans and LLMs raises multiple questions:
Do LLMs, like humans, possesses the ability to choose the most appropriate strategy for solving a given problem~\cite{Wang2024ChainofThoughtRW, yin-etal-2024-reasoning, DBLP:journals/corr/abs-2408-15240, DBLP:journals/corr/abs-2407-21787, zhou2025bridginginternalprobabilityselfconsistency,taubenfeld2025confidenceimprovesselfconsistencyllms}?
Do they, like humans, show a preference for particular strategies~\cite{McCoy2023EmbersOA, shrivastava2025languagemodelspreferknow}?

Prior work by \citet{mondorf-plank-2024-comparing} manually analysed LLM outputs on logical-deduction puzzles and found that each model tends to default to a single preferred strategy, revealing an inherent bias that may limit robustness. They did not, however, test whether different strategies can be invoked on demand or how to combine them effectively.

Building on this observation, this study aims to go further by investigating:
(i) whether an LLM can be explicitly instructed to follow different reasoning strategies,
(ii) whether an LLM can autonomously determine the best strategy for solving a given problem,
and (iii) whether it is possible to guide the model in selecting the most appropriate strategy for a given problem. We believe that answering these questions will not only enable us to make better use of LLMs in reasoning tasks, but also provide deeper insights into their reasoning abilities.

In this paper we present a systematic study of \textbf{strategy-controlled prompting and ensemble selection for LLM reasoning.} We make three contributions:

\begin{itemize}
\item \textbf{Controlled strategy prompting.} We design prompt templates that steer a single LLM into four human-inspired reasoning modes---supposition following, chain construction, compound reasoning, and concatenation---and show in Section~\ref{sec:prompts} that the model adheres to the requested strategy without fine-tuning.

\item \textbf{Empirical analysis of strategy efficacy.} On two logical-deduction benchmarks (\texttt{TruthQuest} and \texttt{ZebraLogic}) we demonstrate in Section~\ref{sec:prompts} that no single strategy dominates. An oracle that always picks the best strategy per problem would raise accuracy by up to 40 percentage points, exposing substantial untapped potential.

\item \textbf{Ensemble-based strategy selection.} Rather than asking the model to choose a strategy, we run all strategies in parallel and select one of the resulting answers using principled combination rules—majority vote, maximum answer probability, minimum entropy, and a model-based verifier. These post-hoc selectors require no meta-prompts or additional training yet consistently outperform any individual strategy prompt as we show in Section~\ref{sec:merge}.
\end{itemize}

The remainder of this article is structured as follows. Section~\ref{sec:tasks} provides a brief introduction to the dataset and the models used in our experiments. In Section~\ref{sec:prompts}, we describe the various reasoning strategies we explore and demonstrate how models can be guided to follow them when solving logical problems. Section~\ref{sec:merge} then explains how these strategies can be combined to enhance performance. Section~\ref{sec:related_work} offers a brief review of related work, and we conclude the article in Section~\ref{sec:conclusion}.

% More precisely, in this work, we show how it's possible by using carefully handcrafted prompts to instruct an LLM to use a predefined strategy. Our experiments, reported in Section~\ref{sec:prompts}, show that while none of the strategies taken individually can improve the overall quality of the model's responses on two corpora containing reasoning tasks, its performance would be significantly better if the model were able to choose the optimal strategy for each problem. In Section~\ref{sec:merge}, we present several methods aimed at enabling the model to automatically determine the optimal strategy.

\section{Experimental Setting \label{sec:tasks}}

All our experiments are conducted within a consistent experimental framework, which we will now briefly outline.

% \paragraph{Datasets}
\subsection{Datasets}
In this work, we investigate the ability of LLMs to solve logical deduction problems—that is, tasks that require systematically deriving a conclusion from a given set of premises. These problems often involve structured reasoning, such as evaluating the validity of an argument, inferring missing information, or detecting contradictions within a logical framework.

In our experiments, we focus on two datasets that have been widely used in prior studies evaluating LLMs' reasoning capabilities: \texttt{TruthQuest}~\cite{mondorf-plank-2024-liar} and \texttt{ZebraLogic}~\cite{lin2025zebralogic}.

\paragraph{\texttt{TruthQuest}} consists of 2,400 questions that require identifying truth-tellers and liars based on their statements. Each instance presents a set of individuals making logical statements about one another, and the goal is to infer who is telling the truth and who is lying. For example, given three individuals (A, B, and C) making the following statements:
\begin{itemize}
\item A: If C is a truth-teller, then B is a liar.
\item B: A is a truth-teller if and only if C is a liar.
\item C: A is a truth-teller.
\end{itemize}
The task is to deduce the correct classification of each individual. The ground truth in this case is that A and C are truth-tellers, while B is a liar.

\paragraph{\texttt{ZebraLogic}}  consists of 1,000 logical puzzles. Unlike \texttt{TruthQuest}, which focuses on binary truth-value assignments, \texttt{ZebraLogic} requires allocating multiple potential values with clues. A logical puzzle consists of $N$ houses numbered increasingly from left to right, each with $M$ distinct attributes (e.g.\ $N$ distinct “person” like Peter and Alice and $N$ distinct “pet” like cat and dog) . Given $K$ clues, the goal is to deduce the unique correct assignment of values to houses. For example, in 2 houses with 2 person and 2 pet mentioned above. Given clues “The person with a cat lives to the left of the person with a dog” and “Alice has a cat”, we can deduce that Peter lives in house 1 with a dog and Alice live in house 2 with a cat.

% \paragraph{Evaluation}
\subsection{Evaluation}
For each problem in our dataset, we prompt an LLM to generate a solution and use a regular expression to check whether the generated text contains the correct answer. To assess the model's ability to reason ---~or more precisely, to produce the correct answer~--- we report accuracy, defined as the percentage of problems for which the generated response includes the correct solution.\footnote{It should be noted that while this evaluation method is commonly used in studies on LLM reasoning capabilities, it does not directly assess the correctness of the model’s reasoning process. Instead, it only verifies whether the final answer is correct, regardless of whether the model arrived at it through valid logical steps, by chance, or via an erroneous reasoning path.}

% \paragraph{Model}
\subsection{Models}
We run our experiments with \texttt{Phi-4-14B}~\cite{Abdin2024Phi4TR}, \texttt{DeepSeek-R1-Distill-Qwen-7B}~\cite{guo2025deepseek}\footnote{We replaced \texttt{DeepSeek-R1-Distill-Qwen-7B} with \texttt{Qwen3-8B} in \texttt{ZebraLogic} due to its poor performance, achieving only around 15\% accuracy in drawing correct conclusions.} and \texttt{Qwen3-8B}~\cite{qwen3}. To ensure comparability of our results, we adopt the same hyperparameter settings as~\newcite{mondorf-plank-2024-comparing}, sampling with top-$p$ set to $0.9$ and the temperature to $0.6$, which encourages more diverse responses than greedy decoding. We will use \texttt{R1-Distill} for short throughout the rest of this article to denote \texttt{DeepSeek-R1-Distill-Qwen-7B}.
%Surely, one might exploit lower top-$p$  and $T$ to have a better reasoning performance~\cite{DBLP:journals/corr/abs-2407-10457}.

\section{Guiding Reasoning Strategies Through Targeted Prompting \label{sec:prompts}}

% \paragraph{Reasonning strategies} 
\subsection{Reasoning strategies}
\newcite{mondorf-plank-2024-comparing} identified four distinct strategies\footnote{\newcite{mondorf-plank-2024-comparing} also introduce the Symbolic Strategy, which focuses more on describing the “format” of the answer rather than a precise reasoning procedure. We could not find a proper way to translate it into inference steps without overlapping with the other strategies, and therefore chose not to include it in our experiments.} that LLMs employ for deductive reasoning problems such as those described in Section~\ref{sec:tasks}:

\begin{itemize}
 \item \textbf{Supposition Following:} Enumerates all propositions, makes a supposition, traces consequences, and tests alternatives if contradictions arise.
 \item \textbf{Chain Construction:} Identifies logical relationships, deduces intermediate implications, and builds a reasoning chain to the conclusion.
 \item \textbf{Compound Strategy:} Integrates multiple logical relationships, iteratively deriving and combining intermediate conclusions.
 \item \textbf{Concatenation Strategy:} Entails the concatenation of two or more statements into a single conclusion that encompasses the logical implications of each combined proposition.
\end{itemize}

\newcite{mondorf-plank-2024-comparing} demonstrated that when LLMs are tasked with solving deductive problems without explicit guidance, each model tends to spontaneously adopt a preferred reasoning strategy. For instance, in their experiments, \texttt{Zephyr-7B-$\beta$} used Supposition Following in 60\% of the cases while \texttt{Llama-2-70B} favored Chain Construction in 50\%. These findings suggest that different LLM architectures might exhibit inherent biases toward specific reasoning pathways.

\subsection{Prompts}
The primary objective of our study is to investigate whether LLMs can be explicitly guided to follow a specified reasoning strategy through targeted prompting. Intuitively, some strategies may be more suitable for particular types of problems, leading to more efficient or direct solutions. To explore this, we designed detailed prompts that explicitly outline each strategy along with its corresponding step-by-step reasoning process.

More precisely, we tested three different methods of specifying the strategy in the prompt:
\begin{enumerate}
    \item providing only a strategy definition;
    \item providing a strategy definition along with a template to complete, such as in the case of Supposition Following:
    \begin{itemize}
        \item Assuming we have a \_.
        \item Then there is a \_.
        \item This means there is no \_.
        \item Thus, there cannot be a \_.
        \item So if \_ then not \_.
        \item \textbf{Answer:} \_
    \end{itemize}
    \item providing a strategy definition and abstract reasoning steps. For instance, Figure~\ref{fig:excerpt_prompt} provides the prompt for \texttt{TruthQuest} used to direct the model towards the Chain Construction strategy (other prompts are provided in the supplementary material). As shown in this example, the structured prompts explicitly guide the model through the expected deductive process while ensuring a systematic approach to reasoning.
\end{enumerate}
For each type of template, we manually reviewed the model's outputs on a few randomly selected examples to assess whether it adhered to the specified strategy. We found that the third formulation produced the best results and therefore adopted it for all subsequent experiments.

% \begin{figure}[ht]
% \centering
% \begin{tcolorbox}[
%     colback=gray!10, 
%     colframe=black, 
%     arc=5mm, 
%     boxrule=0.5pt,
%     boxsep=0mm  % Reduces internal padding
% ]
% \scriptsize
% % \begin{promptbox}
% \tt
% [PROBLEM DESCRIPTION]

% You will reason with chain construction. You construct a chain of propositional statements derived either from the problem description or from intermediate  deductions.

% Let's break it down step by step:

% - Step 1: Identify the logical relationships in each statement, clarifying their conditions.

% - Step 2: Deduce intermediate implications step by step.

% - Step 3: Construct a coherent logical chain and draw a final conclusion in the following format:

% [FORMAT SPECIFICATION AND FINAL INSTRUCTION]
% % \end{promptbox}
% \end{tcolorbox}
% \caption{Excerpt from the prompt guiding the LLM to adopt the Chain Construction strategy.}
% \label{fig:excerpt_prompt}
% \end{figure}

\begin{figure}[ht]
\centering
\begin{tcolorbox}[
    colback=gray!10,
    colframe=black,
    arc=5mm,
    boxrule=0.5pt,
    boxsep=3mm,
    left=1mm, right=1mm, top=1mm, bottom=1mm
]
\scriptsize
\ttfamily
\setstretch{1.2}  % <- this controls the line spacing
[INST] Your task is to solve a logical reasoning problem.\\
You are given a set of statements from which you must logically deduce the identity of a set of characters.\newline
You must infer the identity of each character. First, explain your reasoning. At the end of your answer, you must clearly state the identity of each character by following the format:\newline
Answer:\\
A: ...\\
B: ...\\
C: ...\\
...\newline
\#\#\# Instruction \#\#\#\newline
Assume that there exist only two types of people: knights and knaves. Knights always tell the truth, while knaves always lie.\\
You are given the statements from \{number of characters\} characters. Based on their statements, infer who is a knight and who is a knave.\newline
You will reason with chain construction. You construct a chain of propositional statements derived either from the problem description or from intermediate deductions.\newline
Let’s break it down step by step:\newline
Step 1: Identify the logical relationships in each statement, clarifying their conditions.\newline
Step 2: Deduce intermediate implications step by step based on the statements.\newline
Step 3: Construct a coherent logical chain and draw a final conclusion by following the format:\newline
Answer:\\
A: \{knight/knave\}\\
B: \{knight/knave\}\\
C: \{knight/knave\}\\
...\newline
\#\#\# Now your turn \#\#\#\newline
Based on the following statements, infer who is a knight and who is a knave:\\
\{Question\}\newline
Let’s think step by step. [/INST]
% \end{promptbox}
\end{tcolorbox}
\caption{The prompt guiding the LLM to adopt the Chain Construction strategy for \texttt{TruthQuest}.}
\label{fig:excerpt_prompt}
\end{figure}

% \paragraph{Evaluation}
\subsection{Experimental Results}
\paragraph*{Evaluating Prompt Influence on Strategy Choice} In a first experiment, we evaluate the effectiveness of the different proposed strategies, using the corresponding prompts to query the different models we consider, on the two datasets introduced in Section~\ref{sec:tasks}. 
We aim to evaluate both the model's accuracy in providing the correct answer and its adherence to the strategy specified in the prompt.

First, we checked whether the model followed the strategy suggested by the prompt. For each prompt, we manually annotate answers produced by \texttt{Phi-4-14B} and \texttt{R1-Distill} to 100 randomly selected questions from the \texttt{TruthQuest} dataset. We label each response according to the strategy it follows based on annotation guidelines described in the supplementary material. We also annotate 100 answers that were generated by the prompt of Mondorf and Plank\cite{mondorf-plank-2024-comparing} that does not specify any strategy to reproduce their observations.\footnote{Note that when we used the prompt that does not specify the strategy, some reasoning can rely on multiple strategies which are all counted. That is why the sum of the percentages is greater than 100.}

The annotation results, presented in Table~\ref{tab:anno_both}, allow us to draw two main conclusions. First, we confirm the findings of Mondorf and Plank\cite{mondorf-plank-2024-comparing}: when no specific strategy is provided, the model tends to prefer certain strategies over others. Second, the model generally follows the strategy indicated in the prompt, even though there is some variability in behavior that our preliminary experiments did not manage to explain. This confirms that prompting is an effective way to guide the model toward reasoning strategies it might not ``naturally'' adopt.

\begin{table}[t]
\centering
\resizebox{\columnwidth}{!}{
\begin{tabular}{@{}lcccc@{}}
\toprule
& \multicolumn{2}{c}{Phi-4} &  \multicolumn{2}{c}{R1-Distill}  \\
\cmidrule(lr){2-3} \cmidrule(lr){4-5}
& \makecell{Strategy-\\Specified} & \makecell{No\\Strategy}& \makecell{Strategy-\\Specified} & \makecell{No\\Strategy}\\
\midrule
Supposition Following  & 99\% & 88\% &81\% & 64\% \\
Chain Construction     & 61\% & 12\% &53\% & 11\%\\
Compound Strategy      & 81\% & 12\% &78\% &34\%\\
Concatenation Strategy & 55\% & 17\% &32\% &2\%\\
\bottomrule
\end{tabular}
}
\caption{Percentage of responses by \texttt{Phi-4-14B} and \texttt{R1-Distill} that follow the strategy suggested in the prompt, estimated from a 100 samples per prompt on the \texttt{TruthQuest} dataset.
\label{tab:anno_both}}
\end{table}

\paragraph*{Comparative Performance of Reasoning Strategies} Knowing that prompts can be used to guide the model to follow a specific strategy, we can now evaluate the quality of the responses provided by the two models we consider. Table~\ref{tab:compare_prompt} presents the accuracies achieved using various prompting strategies on our two datasets. It also includes a baseline accuracy (obtained by prompting a model without specifying any strategy) and an \textit{oracle} accuracy, i.e.\ the proportion of problems for which at least one strategy-specific prompt yields the correct answer.  
The gap between the accuracy of a single prompt and that of the corresponding oracle reflects the potential improvement that could be achieved by selecting the appropriate strategy for each problem.

\begin{table}[t]
\centering
\resizebox{\columnwidth}{!}{
\begin{tabular}{lrrcrr}
\toprule
 & \multicolumn{2}{c}{\texttt{TruthQuest}} && \multicolumn{2}{c}{\texttt{ZebraLogic}} \\

\cline{2-3} \cline{5-6}
& Phi-4 & R1-Distill && Phi-4 & Qwen3 \\
\midrule
No strategy        & 47.2\%           & \textbf{63.4\%} && \textbf{27.3\%} & \textbf{32.0\%}\\
\midrule
Supp. Following    & 45.1\%           & 62.7\%          && 26.6\%          &  31.3\%\\
Chain Construction & \textbf{49.0\%}  & 62.5\%          && 25.1\%         & \textbf{32.0\%}\\
Comp. Strategy     & 47.1\%           & 61.4\%          && \underline{27.0\%}          & 31.0\%\\
Concat. Strategy   & \underline{47.4\%} & \underline{63.0\%}        && 25.2\%          &   \underline{31.9\%}\\
\midrule
\textit{Oracle} & \textit{82.9\%}& \textit{90.1\%} && \textit{37.7\%} & \textit{36.4\%}\\
\bottomrule
\end{tabular}
}
\caption{Accuracy of \texttt{Phi4-14B}, \texttt{R1-Distill} and \texttt{Qwen3} on our two datasets for the different prompts we consider. We use \textbf{bold} to denote the best and \underline{underline} to denote the second best}
\label{tab:compare_prompt}
\end{table}

% \textcolor{red}{compare overall performance!!}
We observe that the prompt that specifies no strategy has the best performance across different models and datasets. The chain construction and concatenation strategy performs slightly better than the other strategies.

Our observations indicate that explicitly specifying a strategy in the prompt does not improve problem-solving performance. The five prompts we examined all produced roughly the same results. However, this seemingly negative outcome highlights an important point: when no specific strategy is provided and the model is free to choose its own, its performance is not better than when a strategy is imposed. This suggests that the model is unable to select the best strategy without additional information.

The oracle results further reinforce this conclusion: if the model would be abled evaluate all strategies and choose the most effective one, its responses would be significantly improved. This insight has motivated us to investigate how to combine the outputs of different strategies.

\section{Merging Strategies \label{sec:merge}} 

% \paragraph{Merging criterion}
\subsection{Merging Criteria}

In this section, we explore different ways of combining the predictions of various strategies—an approach that, as shown by the results in the previous section, offers the potential for substantial gains in accuracy. In addition to the widely used majority vote method (denoted \textbf{“\texttt{majority voting}”} in the following), we introduce several new criteria designed to improve overall performance by leveraging statistical confidence measures and model self-evaluation.

\paragraph{Statistical Measures of Confidence} We hypothesize that simple statistical measures can be used to assess the confidence of LLMs in their generated answers. More specifically, we propose two alternative criteria for merging the results produced by the different strategies.

Our first criterion is based on computing the probability of the generated answer as a proxy for model confidence. However, we cannot simply aggregate the probabilities of all generated tokens, since the response consists of two distinct parts: a long reasoning verbalization and a much shorter final answer (e.g., a truth-value assignment in the \texttt{TruthQuest} dataset).

To ensure that the final answer is given appropriate weight, we define the overall response probability as the product of two values: the probability of the reasoning segment and the probability of the answer segment. Each of these is computed as the geometric mean of the token probabilities in its respective part. This criterion is referred to as \textbf{“\texttt{max prob@4}”}.\footnote{More elaborate approaches, such as using a weighted combination of the reasoning and answer probabilities, have not yielded conclusive results. A formal description of these criteria is provided in the supplementary material.}

As an alternative to probability-based confidence, we also consider the model’s uncertainty by measuring the mean entropy of the generated response. This criterion, denoted \textbf{“\texttt{min entropy@4}”}, selects the response with the lowest average entropy across its tokens. Lower entropy indicates higher model confidence, while higher entropy suggests greater uncertainty and potential variability in the output.

\paragraph{Model-based Assessment} We also introduce a more sophisticated criterion, denoted \textbf{“\texttt{verifier}”}, which relies on an “external” LLM to assess the soundness of a reasoning. In this approach, the answer for each prompt is split into approximately 100-word chunks, truncated by sentence boundaries. \texttt{R1-Distill} is then prompted\footnote{Assume we have $n$ chunks, the $i$-th prompt that we use is:
\begin{Verbatim}
You are a reasoning assistant. Your job is to determine
whether the answer is logically valid.

Question: {question}
Answer: {chunk 0} ... {chunk i}

Is the reasoning correct so far?
My answer is (Yes or No):
\end{Verbatim}
} to verify the correctness of each chunk. A self-confidence score for each chunk is defined as the probability that the model generates “Yes” in response to this verification prompt. The overall self-confidence score for an answer is then computed by averaging the self-confidence scores of its individual chunks, and the answer with the highest average score is selected. For responses generated by \texttt{Phi-4} in the \texttt{ZebraLogic} dataset, there are 7.54 chunks on average per response, with a range from 2 to 18.

\paragraph{Combined Merging Criteria} Finally, we consider two hybrid approaches: \textbf{\texttt{vote + prob}} and \textbf{\texttt{vote + verifier}}. Both begin by selecting the majority answer across strategies. In case of a tie, the final answer is chosen based on an auxiliary signal: accumulated response probability for \texttt{vote + prob}, and verification probability for \texttt{vote + verifier}. In the \texttt{ZebraLogic} dataset, 2.6\% of the questions answered by \texttt{Phi-4} result in a tie in majority voting, thus requiring disambiguation via these secondary criteria.

\subsection{Experimental Results}

Table~\ref{tab:overall} presents the results of the merging strategies described in the previous section. The results reveal that no single merging strategy consistently outperforms the others: Depending on the model and dataset under consideration, different strategies achieve the best performance. This variability suggests that the effectiveness of a merging strategy is influenced by both the underlying model architecture and the nature of the task.

Notably, the \texttt{ZebraLogic} dataset yields consistently lower performance gains from merging strategies. This outcome may be attributed to the dataset's more complex, multi-valued reasoning tasks, which appear less amenable to the benefits of strategy combination. In contrast, the \texttt{TruthQuest} dataset shows more substantial improvements, indicating that it benefits more from the integration of diverse reasoning strategies.

Nevertheless, several trends can be identified across datasets. The \texttt{majority vote} strategy tends to outperform the pure-probability-based approaches (\texttt{min entropy@4} and \texttt{max prob@4}). This finding suggests that consensus among different reasoning strategies may serve as a more reliable signal of correctness than token-level confidence alone.

The \texttt{verifier}-based method exceeds the performance of majority vote in most cases. However, it requires additional decoding steps, which introduces non-negligible computational overhead. Given the relatively modest improvements in accuracy, the cost of inference may outweigh the benefits in certain applications.

Hybrid strategies that combine two criteria offer a potential solution to this problem. In particular, \texttt{vote + verifier} selectively invokes additional decoding only when necessary to resolve ties, thereby reducing resource usage. This strategy demonstrates strong performance, frequently ranking as the best or second-best method.

Despite these advances, all merging strategies evaluated fall short of the oracle baseline, highlighting that current criteria do not fully capitalize on the potential to select the optimal reasoning strategy across all instances.

%The oracle score is the highest score we could achieve by combining different strategies. The higher it is, the easier it is to find a correct answer from another strategy when one strategy makes a mistake. In other words, the higher it is, the higher the boost you can generally achieve by combining different strategies. For example, \texttt{R1-Distill} shows the highest oracle, the best merging method \emph{vote + prob} is 9.2\%(14.5\% relatively) higher than prompt with no strategy specified. In contrast, \texttt{Qwen3} has the lowest oracle, and its best merging method \emph{min entropy@4} is only 2.4\% (7.5\% relatively) higher than prompt without strategy.

\begin{table}
\centering
\resizebox{\columnwidth}{!}{
\begin{tabular}{lrrrr}
\toprule
 & \multicolumn{2}{c}{\texttt{TruthQuest}}& \multicolumn{2}{c}{\texttt{ZebraLogic}}\\
\cline{2-3} \cline{4-5}
& Phi-4& R1-Distill& Phi-4& Qwen3\\
\midrule
best single strategy & 49.0\%& 63.4\%& 27.3\%& 32.0\%\\
\midrule
majority vote   & 54.9\%             &68.4\%              & 27.9\%         &32.0\%\\
\midrule
min entropy@4   & 48.8\%             &67.0\%              & 27.1\%         &\textbf{34.4\%}\\
max prob@4      & 48.4\%             &64.7\%              & 26.5\%         &\underline{34.3\%}\\
verifier        & 46.9\%             &\textbf{74.1\%}     & 28.2\%         &33.4\% \\
\midrule
vote + prob     &\textbf{57.3\%}     &69.7\%              &\underline{28.2\%}          &32.9\%\\
vote + verifier &\underline{55.8\%}  &\underline{72.6\%}  &\textbf{28.3\%} &32.8\%\\
\midrule
\textit{Oracle} & \textit{82.9\%}& \textit{90.1\%} & \textit{37.7\%} & \textit{36.4\%} \\
\bottomrule
\end{tabular}
}
\caption{Accuracy (in \%) of our four merging criteria. See Table~\ref{tab:compare_prompt} for results of the base strategies. We use \textbf{bold} to denote the best and \underline{underline} to denote the second best.\label{tab:overall}}
\end{table}

\subsection{Analysis of Accuracy by Problem Difficulty \label{overall_set}}

\paragraph{Defining Problem Difficulty} To gain deeper insight into the results described in the previous section and to better understand the limitations of the various merging strategies ---~specifically, when they succeed and when they fail~--- we also analyzed performance as a function of problem difficulty. For both datasets considered, there exist natural ways to quantify the difficulty of individual problems.

In the case of \texttt{TruthQuest}, we used the number of characters involved in a problem as a proxy for its complexity. Intuitively, a problem with fewer characters ---~and therefore fewer variables~--- is easier to solve than one with more. This is based on the idea that increasing the number of characters leads to more logical relationships to analyze. The dataset includes problems with between 3 and 6 characters, allowing us to define four levels of increasing difficulty based on this criterion.

For \texttt{ZebraLogic}, we adopted the difficulty criteria defined in~\cite{zebralogic2024}. Specifically, questions involving smaller configurations ---~namely 2 houses by 2, 3, 4, 5, or 6 features, as well as 3 houses by 2 or 3 features~--- are classified as \textit{easy}. The remaining 18 larger-sized configurations are considered \textit{harder} questions.

\begin{table}[t]
\centering
\resizebox{\columnwidth}{!}{
\begin{tabular}{llccccc}
\toprule
& \textbf{Complexity}& \textbf{3 Person} & \textbf{4 Person} & \textbf{5 Person} & \textbf{6 Person} & \textbf{Avg.} \\
\midrule
\multicolumn{7}{l}{\textit{Single strategy}} \\
& No strategy       & \underline{65.7\%} & 49.7\% & 40.5\% & 33.5\% & 47.2\% \\
& Supposition       & 60.5\% & 49.7\% & 39.3\% & 30.8\% & 45.1\% \\
& Chain             & \textbf{66.7\%} & \textbf{50.5\%} & \textbf{44.7\%} & \underline{35.2\%} & \textbf{49.0\%} \\
& Compound          & 63.2\% & \underline{50.2\%} & 39.7\% & 35.3\% & 47.1\% \\
& Concatenation     & 61.7\% & 49.2\% & \underline{42.5\%} & \textbf{36.3\%} & \underline{47.4\%} \\
\midrule
\multicolumn{7}{l}{\textit{Combined strategies}} \\
& min entropy@4     & 65.3\% & 50.3\% & 43.3\% & 36.5\% & 48.8\% \\
& max prob@4        & 63.7\% & 50.2\% & 42.7\% & 37.3\% & 48.4\% \\
& verifier          & 61.5\% & 47.5\% & 44.7\% & 34.0\% & 46.9\% \\
& majority vote     & 69.0\% & 57.5\% & \underline{50.5\%} & 42.7\% & 54.9\% \\
& vote + prob       & \textbf{72.2\%} & \textbf{60.5\%} & \textbf{51.2\%} & \textbf{45.2\%} & \textbf{57.3\%} \\
& vote + verifier   & \underline{70.2\%} & \underline{59.7\%} & 50.3\% & \underline{42.8\%} & \underline{55.8\%} \\
\midrule
\multicolumn{2}{l}{Oracle}   & \textit{91.8\%} & \textit{87.3\%} & \textit{78.5\%} & \textit{73.8\%} & 82.9\% \\
\bottomrule
\end{tabular}
}
\caption{Overall accuracy of \texttt{Phi-4} across different test subsets in \texttt{TruthQuest}.}
\label{overall_set_phi}
\end{table}

\begin{table}[t]
\centering
\resizebox{\columnwidth}{!}{
\begin{tabular}{llccccc}
\toprule
& \textbf{Complexity} & \textbf{3 Person} & \textbf{4 Person} & \textbf{5 Person} & \textbf{6 Person} & \textbf{Avg.} \\
\midrule
\multicolumn{7}{l}{\textit{Single strategy}} \\
& No strategy       & 74.2\% & \textbf{69.5\%} & \textbf{61.0\%} & \textbf{49.0\%} & \textbf{63.4\%} \\
& Supposition       & 75.3\% & \underline{68.3\%} & 59.3\% & \underline{47.8\%} & 62.7\% \\
& Chain             & \underline{77.7\%} & 66.3\% & 60.0\% & 45.8\% & 62.5\% \\
& Compound          & 76.0\% & 67.3\% & 55.0\% & 47.3\% & 61.4\% \\
& Concatenation     & \textbf{79.3\%} & 68.2\% & \underline{60.7\%} & 44.0\% & \underline{63.0\%} \\
\midrule
\multicolumn{7}{l}{\textit{Combined strategies}} \\
& min entropy@4     & 79.2\% & 70.5\% & 64.2\% & 54.3\% & 67.0\% \\
& max prob@4        & 78.8\% & 70.2\% & 60.0\% & 49.8\% & 64.7\% \\
& verifier          & \textbf{88.8\%} & \underline{78.5\%} & \underline{70.2\%} & \textbf{58.8\%} & \textbf{74.1\%} \\
& majority vote     & 81.0\% & 75.1\% & 67.0\% & 50.6\% & 68.4\% \\
& vote + prob       & 82.0\% & 75.8\% & 68.7\% & 52.3\% & 69.7\% \\
& vote + verifier   & \underline{85.7\%} & \textbf{78.7\%} & \textbf{70.7\%} & \underline{55.3\%} & \underline{72.6\%} \\
\midrule
\multicolumn{2}{l}{\textit{Oracle}}   & \textit{97.3\%} & \textit{93.0\%} & \textit{89.5\%} & \textit{80.5\%} & 90.1\% \\
\bottomrule
\end{tabular}
}
\caption{Overall accuracy of \texttt{R1-Distill} across different test subsets in \texttt{TruthQuest}.}
\label{overall_set_deepseek}
\end{table}

\begin{table}[t]
  \centering
  \resizebox{\columnwidth}{!}{
  \begin{tabular}{llccc}
    \toprule
    & \textbf{Complexity} & \textbf{Easy Avg.} & \textbf{Hard Avg.} & \textbf{All Avg.} \\
    \midrule
    \multicolumn{5}{l}{\textit{Single strategy}} \\
    & No strategy       & 58.6\%  & \textbf{15.1\%} & \textbf{27.3\%} \\
    & Supposition       & 62.9\%  & 12.5\%  & 26.6\% \\
    & Chain             & \textbf{65.3\%} & 9.4\%   & 25.1\% \\
    & Compound          & 63.5\%  & \underline{12.7\%} & \underline{27.0\%} \\
    & Concatenation     & \underline{65.0\%} & 9.7\%   & 25.2\% \\
    \multicolumn{5}{l}{\textit{Combined strategies}} \\
    & min entropy@4     & 65.7\%  & \underline{12.5\%} & 27.1\% \\
    & max prob@4        & 65.3\%  & 11.4\%  & 26.5\% \\
    & verifier          & \textbf{68.9\%} & 12.4\%  & \underline{28.2\%} \\
    & majority vote     & \underline{67.0\%} & \textbf{12.6\%} & 27.9\% \\
    & vote + prob       & \textbf{68.9\%} & 12.4\%  & \underline{28.2\%} \\
    & vote + verifier   & \textbf{68.9\%} & \underline{12.5\%} & \textbf{28.3\%} \\
    \midrule
    \multicolumn{2}{l}{\textit{Oracle}}  & \textit{76.8\%} & \textit{22.5\%} & \textit{37.7\%} \\
    \bottomrule
  \end{tabular}
  }
  \caption{Overall accuracy of \texttt{Phi-4} across different subsets in \texttt{ZebraLogic}.}
  \label{overall_set_zebra}
\end{table}

\begin{table}[t]
  \centering
  \resizebox{\columnwidth}{!}{
  \begin{tabular}{llccc}
    \toprule
    & \textbf{Complexity} & \textbf{Easy Avg.} & \textbf{Hard Avg.} & \textbf{All Avg.} \\
    \midrule
    \multicolumn{5}{l}{\textit{Single strategy}} \\
&No strategy      & \textbf{93.5\%}  & 8.0\%   & \textbf{32.0\%}\\
&Supposition      & 92.5\%  & 7.5\%   & 31.3\% \\
&Chain            & 91.1\%  & \textbf{9.0\%}   & \textbf{32.0\%}\\
&Compound         & 90.3\%  & 7.9\%   & 31.0\% \\
&Concatenation    & \underline{93.2\%}& \underline{8.1\%}& \underline{31.9\%}\\
\multicolumn{5}{l}{\textit{Combined strategies}} \\
&min entropy@4    & \textbf{93.6\%} & \textbf{11.4\%} & \textbf{34.4\%} \\
&max prob@4       & \underline{93.2\%}& \textbf{11.4\%}& \underline{34.3\%}\\
% verifier(1.5B)&  & &  \\
&verifier         & 93.2\%  & \underline{10.1\%}& 33.4\% \\
&majority vote    & 92.2\%  & 8.5\%   & 32.0\% \\
&vote + prob      & \underline{93.2\%}& 9.4\%   & 32.9\% \\
% vote + verifier(1.5B) & 92.9 & 9.2 & 32.6 \\
&vote + verifier  & 92.9\%  & 9.4\%   & 32.8\% \\
\midrule
\multicolumn{2}{l}{Oracle}  & 95.3\%  & 13.4\%  & 36.4\% \\
     \bottomrule
  \end{tabular}
  }
   \caption{Overall accuracy of \texttt{Qwen3-8B} across different subsets in \texttt{ZebraLogic}.} \label{overall_set_zebra_qwen3}
\end{table}

The results for the \texttt{TruthQuest} benchmark are detailed in Tables \ref{overall_set_phi} and \ref{overall_set_deepseek}, and for \texttt{ZebraLogic} in Tables \ref{overall_set_zebra} and \ref{overall_set_zebra_qwen3}. These tables present the performance across various configurations and allow for a fine-grained analysis of how different approaches behave depending on the complexity of the problems.

One key observation that emerges from the oracle scores is that they decrease consistently as the complexity of the problems increases. This behavior aligns with our expectations and provides empirical validation for our method of defining problem complexity—based, in this case, on the number of characters involved in the logical puzzle.

As observed in our preliminary experiments, the results remain highly dependent on the specific employed model. Different models yield varying levels of performance, highlighting the importance of model capabilities in tasks that require structured reasoning.

Despite this variability, a general pattern can be identified. Verifier-based approaches—those that explicitly assess the correctness or soundness of intermediate reasoning steps—tend to produce the most substantial improvements in simpler problem settings. However, as problem complexity increases, these approaches often fail to provide significant benefits. Instead, more flexible strategies that do not enforce a specific reasoning method, and that rely solely on statistical combination techniques (e.g., majority voting or confidence-weighted scoring), tend to perform better on the more difficult problems.

This suggests a limitation in the current verifier models: they seem to be effective only for evaluating short or straightforward chains of reasoning. When confronted with more complex or longer reasoning processes, the verifier’s ability to accurately assess soundness diminishes, which in turn limits its usefulness in guiding the model toward correct final answers.

An additional factor that may contribute to this effect is that the oracle gains—i.e., the theoretical maximum performance improvement achievable through optimal selection—are lower in more complex tasks. As a result, even when improved selection methods are used, the absolute gains remain modest and harder to realize in practice.

\section{Related Work\label{sec:related_work}}
\paragraph{Prompting Methods for LLM Reasoning:}
The discovery of chain-of-thought prompting has spurred substantial research into improving LLM reasoning via prompt design. In their seminal work, \citet{Wei2022ChainOT}
showed that providing examples of step-by-step reasoning can unlock latent reasoning capabilities in large models, achieving impressive results on math and logic problems.
\citet{Kojima_Gu_Reid_Matsuo_Iwasawa_2022}
later found that even without examples, simply appending a generic trigger phrase (e.g., ``Let's think step by step'') allows LLMs to perform complex reasoning zero-shot, dramatically improving accuracy on benchmarks. Building on these ideas, researchers have proposed more sophisticated prompting strategies to handle harder tasks. For instance, least-to-most prompting decomposes a complex problem into a sequence of simpler sub-problems that the model solves one by one \cite{Zhou23LeastToMost}. This approach enables better generalization to difficult questions, outperforming standard chain-of-thought by a wide margin on compositional tasks (e.g., 99\% vs 16\% accuracy on the SCAN challenge).
Other methods include prompts that encourage planning or tool use (e.g., generating code or queries to external knowledge bases) to assist reasoning \cite{Wang2023SelfConsistency}. These works demonstrate that prompt engineering can significantly influence an LLM's reasoning process. However, they typically assume a \textit{fixed reasoning format} (be it a chain-of-thought, least-to-most breakdown, or code-generation approach) applied uniformly to all inputs.
In contrast, our work treats reasoning style as a conditional choice: we explicitly prompt the model with different strategies and show the benefits of selecting among them adaptively. To the best of our knowledge, a systematic exploration of \textit{multiple distinct reasoning strategies within the same model} has not been addressed in prior prompting research.

\paragraph{Multiple Reasoning Paths and Self-Consistency:}
Even when using a fixed strategy, recent studies have noted that a given problem might be solved via different reasoning paths.
\citet{Wang2023SelfConsistency} introduced a self-consistency decoding strategy that samples diverse chains-of-thought and then selects the answer most consistent across these different reasoning trials. This method boosted reasoning accuracy by ensembling the model’s reasoning outcomes, \textit{implicitly acknowledging that multiple approaches can reach a correct answer}.
Similarly, methods like self-refinement and debate prompts have attempted to have the model generate and evaluate multiple solution paths before finalizing an answer, in order to increase reliability. These approaches, however, differ from our aim: rather than sampling stochastically varied reasoning traces, \textit{we directly control the reasoning strategy} the model employs.
Our strategy-conditioned prompting could be seen as orthogonal to self-consistency --- in fact, one could combine them by sampling each distinct strategy for a given problem and then choosing the most confident result. We leave such combinations to future work, and focus here on demonstrating the core effect of strategy guidance and selection.

\paragraph{Reasoning Strategy Biases and Adaptivity:}
Our work is inspired by analyses of how LLMs reason in the absence of explicit instructions. Beyond the chain-of-thought successes, researchers have begun to ask whether models exhibit inherent reasoning preferences.
\citet{mondorf-plank-2024-comparing} provided evidence that different LLMs gravitate toward particular solution strategies on logical deduction tasks.
This finding suggests that factors like pre-training data or model architecture could bias a model's reasoning style, but it left open the question of whether such style can be changed or optimized. Some recent efforts have started to explore \textit{adaptive reasoning}.
For example, \citet{Zhou23LeastToMost} hinted at strategy selection by choosing between letting the model solve a problem directly vs. breaking it into parts, depending on the problem's perceived difficulty.
More directly, \citet{xu2025teachingllmsaccordingaptitude} propose a training-time framework that allows an LLM to \textit{learn} when to apply chain-of-thought reasoning versus a calculator-like tool, effectively personalizing the strategy to the model's strengths.
These approaches either require additional training/fine-tuning or focus on narrow cases (e.g. math problems only). In contrast, we tackle strategy adaptivity at inference time on general logical problems, using prompting techniques that work with off-the-shelf models. To our knowledge, our study is the first to demonstrate that an LLM can be prompted to switch among multiple reasoning strategies and that doing so yields performance gains on challenging NLP tasks.

\section{Conclusion\label{sec:conclusion}}
We presented a systematic exploration of strategy-controlled prompting combined with post-hoc ensemble selection for logical reasoning in large language models. By crafting prompts that explicitly invoke four distinct human-inspired reasoning strategies, we showed that an off-the-shelf LLM can be steered into different reasoning modes without fine-tuning. Extensive experiments on the \texttt{TruthQuest} and \texttt{ZebraLogic} benchmarks revealed that each strategy has complementary strengths, producing an oracle gap of up to 40 points between the best-choice-per-instance and any single fixed strategy.

To exploit this diversity, we ran all strategies in parallel and selected an answer using lightweight combination methods. These selectors, which require no additional training or meta-prompting, lifted accuracy by 7–11 points on \texttt{TruthQuest} and 1–3 points on \texttt{ZebraLogic}, consistently outperforming every individual strategy prompt. Our findings demonstrate that reasoning style is a controllable latent variable and that simple ensemble methods can substantially enhance LLM robustness, offering an effective and practical path toward more human-like, adaptable problem-solving with current models.

\section*{Acknowledgements}
This work is partially supported by a public grant overseen by the French National Research Agency (ANR) as part of the program Investissements d’Avenir (ANR-10-LABX-0083).

\bibliography{custom}

\clearpage

\appendix

\section{Annotation Rules} \label{anno_rule}
Annotation rules are used to identify the strategy in a model's response.
They are necessary but insufficient conditions for each strategy.
% We make our annotation mostly based on this rule, but not all of them.
\begin{itemize}
    \item Supposition Following: If a response makes an assumption and then evaluates whether a given statement is consistent or contradictory to that assumption, it is classified as using the supposition following strategy. Keywords such as "suppose" and "assume" impose a high probability of following the strategy.
    \item Chain Construction: If a response follows a step-by-step reasoning process where each step logically leads to the next until reaching a conclusion, it is labeled as following the chain construction strategy.
    \item Compound Strategy: If a response derives intermediate conclusions before reaching the final conclusion, it is categorized as following the compound strategy.
    \item Concatenation Strategy: If a response concatenate two or more statements into a single conclusion, it is classified as using the concatenation strategy. Keywords such as "link" and "concatenate" impose a high probability of following the strategy.

\end{itemize}

While most of the responses did contains a bit of suppositional assumption in certain steps. If it is only a tactical step without further expansion, we will not annotate it to suppositional following strategy.

\section{Response Probability \label{sec:proba}}
For merging strategies, we used an approximation of the conditional probability of the rational sequence $\vr$ given question $\vq$, strategy description $\vs$ and formating instructions $\vx$. For this, we consider a token-wise decomposition. Each token in $\vr$ is generated based on prior tokens. We use geometric mean to make samples of different lengths comparable.

\[
\begin{aligned}
P_{\text{rational}} &= P(\vr \mid \vq, \vs, \vx) \\
&= \prod_{n=0}^{N} P\Bigl(r_n \,\big|\, \vr_{0:n-1},\, \vq,\, \vs,\, \vx\Bigr) \\
&\propto \exp\Biggl( \frac{1}{N+1} \sum_{i=0}^{N} \log P\Bigl(r_i \,\big|\, \vr_{0:i-1},\, \vq,\, \vs,\, \vx\Bigr) \Biggr)
\end{aligned}
\]

where:
\begin{itemize}
   \item $\vr = (r_0, \dots, r_N)$ represents the output sequence of tokens in the rational
   \item $\vr_{0:n-1}$ denotes $(r_0, \dots, r_{n-1})$
\end{itemize}

Apart from that, we also care about whether the final answer sequence $\va$ is correct or not. The definition is similar to the way we define rational probability:

\[
\begin{aligned}
P_{\text{answer}} &= P\Bigl(\va \mid \vq, \vr, \vs, \vx\Bigr) \\
&= \prod_{j=0}^{M} P\Bigl(a_j \,\big|\, \va_{0:j-1},\, \vq,\, \vr,\, \vs,\, \vx\Bigr) \\
&\propto \exp\!\Biggl( \frac{1}{M+1}\\
&\sum_{m=0}^{M} \log P\Bigl(a_m \,\big|\, a_{0:m-1},\, q,\, \vr,\, s,\, x\Bigr) \Biggr)
\end{aligned}
\]

where:
\begin{itemize}
   \item $\va = (a_0, \dots, a_M)$ represents the output sequence of tokens in rational
   \item $\va_{0:j-1}$ denotes $(a_0, \dots, a_{j-1})$
\end{itemize}

To combine them into one single metric, we multiple them together because it is similar to the way that we calculate the total probability of the response.

\[
P_{\text{combined}} =  {P_{\text{rational}}}  \times {P_{\text{answer}}}
\]

We show the result of probability distribution and their correctness for the first 20 problems in \texttt{TruthQuest} dataset in Table \ref{prob}.

\begin{table}[t]
    \centering
    \renewcommand{\arraystretch}{1.2}
    \resizebox{\columnwidth}{!}{
    \begin{tabular}{lccccc}
        % \toprule
        \rotatebox{65}{\textbf{Problem N$^\circ$}}
        & \rotatebox{65}{\textbf{No Strategy}} 
        & \rotatebox{65}{\textbf{Supposition}} 
        & \rotatebox{65}{\textbf{Chain}} 
        & \rotatebox{65}{\textbf{Compound}} 
        & \rotatebox{65}{\textbf{Concatenation}} \\
        \midrule
        1  & \cellcolor{correct}\textbf{0.238} & \cellcolor{correct}0.209 & \cellcolor{correct}0.230 & \cellcolor{correct}0.222 & \cellcolor{correct}0.227 \\
        2  & \cellcolor{incorrect}0.237 & \cellcolor{correct}0.211 & \cellcolor{correct}0.212 & \cellcolor{correct}\textbf{0.240} & \cellcolor{correct}0.211 \\
        3  & \cellcolor{correct}\textbf{0.234} & \cellcolor{incorrect}0.218 & \cellcolor{correct}0.212 & \cellcolor{correct}0.225 & \cellcolor{correct}0.210 \\
        4  & \cellcolor{correct}\textbf{0.236} & \cellcolor{correct}0.205 & \cellcolor{correct}0.221 & \cellcolor{correct}0.202 & \cellcolor{correct}0.226 \\
        5  & \cellcolor{correct}\textbf{0.240} & \cellcolor{correct}0.226 & \cellcolor{correct}0.229 & \cellcolor{correct}0.234 & \cellcolor{correct}0.224 \\
        6  & \cellcolor{correct}0.238 & \cellcolor{incorrect}0.199 & \cellcolor{correct}\textbf{0.239} & \cellcolor{incorrect}0.217 & \cellcolor{incorrect}0.216 \\
        7  & \cellcolor{incorrect}\textbf{0.233} & \cellcolor{incorrect}0.224 & \cellcolor{incorrect}0.188 & \cellcolor{incorrect}0.229 & \cellcolor{incorrect}0.227 \\
        8  & \cellcolor{correct}\textbf{0.238} & \cellcolor{incorrect}0.220 & \cellcolor{correct}0.233 & \cellcolor{correct}0.221 & \cellcolor{correct}0.227 \\
        9  & \cellcolor{correct}\textbf{0.235} & \cellcolor{correct}0.224 & \cellcolor{correct}0.231 & \cellcolor{correct}0.202 & \cellcolor{correct}0.227 \\
        10 & \cellcolor{incorrect}\textbf{0.237} & \cellcolor{correct}0.236 & \cellcolor{incorrect}0.217 & \cellcolor{correct}0.223 & \cellcolor{correct}0.233 \\
        11 & \cellcolor{correct}\textbf{0.237} & \cellcolor{incorrect}0.232 & \cellcolor{correct}\textbf{0.237} & \cellcolor{correct}0.204 & \cellcolor{correct}\textit{Nan} \\
        12 & \cellcolor{incorrect}\textbf{0.235} & \cellcolor{incorrect}0.229 & \cellcolor{incorrect}0.225 & \cellcolor{incorrect}0.229 & \cellcolor{correct}0.218 \\
        13 & \cellcolor{correct}\textbf{0.236} & \cellcolor{correct}0.214 & \cellcolor{correct}0.220 & \cellcolor{correct}0.221 & \cellcolor{correct}0.223 \\
        14 & \cellcolor{incorrect}\textbf{0.234} & \cellcolor{correct}0.230 & \cellcolor{correct}0.207 & \cellcolor{correct}0.217 & \cellcolor{correct}0.226 \\
        15 & \cellcolor{correct}\textbf{0.240} & \cellcolor{correct}0.226 & \cellcolor{correct}0.237 & \cellcolor{correct}0.229 & \cellcolor{correct}0.210 \\
        16 & \cellcolor{correct}\textbf{0.242} & \cellcolor{incorrect}0.218 & \cellcolor{correct}0.226 & \cellcolor{correct}0.233 & \cellcolor{correct}0.231 \\
        17 & \cellcolor{correct}\textbf{0.236} & \cellcolor{incorrect}0.220 & \cellcolor{incorrect}0.206 & \cellcolor{correct}0.216 & \cellcolor{correct}0.224 \\
        18 & \cellcolor{correct}0.233 & \cellcolor{correct}0.225 & \cellcolor{correct}\textbf{0.235} & \cellcolor{correct}0.202 & \cellcolor{correct}0.226 \\
        19 & \cellcolor{correct}\textbf{0.236} & \cellcolor{correct}0.211 & \cellcolor{correct}0.231 & \cellcolor{correct}0.209 & \cellcolor{correct}0.221 \\
        20 & \cellcolor{incorrect}\textbf{0.237} & \cellcolor{correct}0.218 & \cellcolor{incorrect}0.223 & \cellcolor{incorrect}0.231 & \cellcolor{correct}0.235 \\
        \bottomrule
    \end{tabular}
    }
    \caption{The value of $P_{\text{combined}}$ of Phi-4 responses for different strategies for the first 20 problems from \texttt{TruthQuest}. Maximum probability is in bold. Green denotes that the conclusion is valid, red denotes invalid. \textit{Nan} denotes that we did not successfully parse the key word ("Answser:") at the end of the response.}
    \label{prob}
\end{table}

\section{Entropy Computation \label{entropy_cal}}

For each token  $r_i$ , the entropy is defined as:

\begin{multline*}
H_i = - \sum_{k \in \gV} P(r_i = k \mid \vr_{0:i-1}, \vq, \vs, \vx) \\
\times \log P(r_i = k \mid \vr_{0:i-1}, \vq, \vs, \vx)
\end{multline*}

where $\gV$ is the vocabulary.

To calculate the overall uncertainty for the rational sequence $\vr$, the average entropy is:

\[
H_{\text{rational}} = \frac{1}{N+1} \sum_{i=0}^{N} H_i
\]

Similarly, for the final answer $\va$, the entropy is defined as:

\begin{multline*}
H_m = - \sum_{j \in \gV} P(a_m = j \mid \va_{0:m-1}, \vq, \vr, \vs, \vx) \\
\times \log P(a_m = j \mid \va_{0:m-1}, \vq, \vr, \vs, \vx)
\end{multline*}

where $\gV$ is the vocabulary size.

The average entropy for the answer sequence is calculated as:

\[
H_{\text{answer}} = \frac{1}{M+1} \sum_{m=0}^{M} H_m
\]

We use the arithmetic average to combine the rational entropy and the answer entropy into a single score.

$$H_{\text{combined}} = \frac{1}{2} H_{\text{rational}} + \frac{1}{2} H_{\text{answer}}$$

We show the entropy distribution with first 20 questions of \texttt{TruthQuest} in Table \ref{ent}. 

\begin{table}[t]
    \centering
    \renewcommand{\arraystretch}{1.2}
    \resizebox{\columnwidth}{!}{
    \begin{tabular}{lccccc}
        \rotatebox{65}{\textbf{Problem N$^\circ$}}
        & \rotatebox{65}{\textbf{No Strategy}} 
        & \rotatebox{65}{\textbf{Supposition}} 
        & \rotatebox{65}{\textbf{Chain}} 
        & \rotatebox{65}{\textbf{Compound}} 
        & \rotatebox{65}{\textbf{Concatenation}} \\
        \midrule
        1  & \cellcolor{correct}\textbf{0.044} & \cellcolor{correct}0.110 & \cellcolor{correct}0.078 & \cellcolor{correct}0.075 & \cellcolor{correct}0.075 \\
        2  & \cellcolor{incorrect}\textbf{0.041} & \cellcolor{correct}0.055 & \cellcolor{correct}0.107 & \cellcolor{correct}0.049 & \cellcolor{correct}0.102 \\
        3  & \cellcolor{correct}\textbf{0.052} & \cellcolor{incorrect}0.119 & \cellcolor{correct}0.095 & \cellcolor{correct}0.100 & \cellcolor{correct}0.102 \\
        4  & \cellcolor{correct}\textbf{0.039} & \cellcolor{correct}0.139 & \cellcolor{correct}0.101 & \cellcolor{correct}0.133 & \cellcolor{correct}0.080 \\
        5  & \cellcolor{correct}\textbf{0.034} & \cellcolor{correct}0.060 & \cellcolor{correct}0.067 & \cellcolor{correct}0.044 & \cellcolor{correct}0.079 \\
        6  & \cellcolor{correct}0.041 & \cellcolor{incorrect}0.115 & \cellcolor{correct}\textbf{0.035} & \cellcolor{incorrect}0.069 & \cellcolor{incorrect}0.090 \\
        7  & \cellcolor{incorrect}\textbf{0.051} & \cellcolor{incorrect}0.092 & \cellcolor{incorrect}0.127 & \cellcolor{incorrect}0.078 & \cellcolor{incorrect}0.075 \\
        8  & \cellcolor{correct}\textbf{0.038} & \cellcolor{incorrect}0.104 & \cellcolor{correct}0.066 & \cellcolor{correct}0.102 & \cellcolor{correct}0.074 \\
        9  & \cellcolor{correct}\textbf{0.043} & \cellcolor{correct}0.062 & \cellcolor{correct}0.072 & \cellcolor{correct}0.140 & \cellcolor{correct}0.081 \\
        10 & \cellcolor{incorrect}\textbf{0.042} & \cellcolor{correct}0.056 & \cellcolor{incorrect}0.124 & \cellcolor{correct}0.101 & \cellcolor{correct}0.069 \\
        11 & \cellcolor{correct}0.046 & \cellcolor{incorrect}0.071 & \cellcolor{correct}\textbf{0.042} & \cellcolor{correct}0.084 & \cellcolor{correct}\textit{Nan} \\
        12 & \cellcolor{incorrect}\textbf{0.045} & \cellcolor{incorrect}0.073 & \cellcolor{incorrect}0.109 & \cellcolor{incorrect}0.067 & \cellcolor{correct}0.086 \\
        13 & \cellcolor{correct}\textbf{0.038} & \cellcolor{correct}0.118 & \cellcolor{correct}0.108 & \cellcolor{correct}0.104 & \cellcolor{correct}0.070 \\
        14 & \cellcolor{incorrect}\textbf{0.041} & \cellcolor{correct}0.053 & \cellcolor{correct}0.057 & \cellcolor{correct}0.110 & \cellcolor{correct}0.097 \\
        15 & \cellcolor{correct}\textbf{0.031} & \cellcolor{correct}0.066 & \cellcolor{correct}0.059 & \cellcolor{correct}0.068 & \cellcolor{correct}0.099 \\
        16 & \cellcolor{correct}\textbf{0.023} & \cellcolor{incorrect}0.129 & \cellcolor{correct}0.090 & \cellcolor{correct}0.064 & \cellcolor{correct}0.069 \\
        17 & \cellcolor{correct}\textbf{0.036} & \cellcolor{incorrect}0.121 & \cellcolor{incorrect}0.081 & \cellcolor{correct}0.114 & \cellcolor{correct}0.079 \\
        18 & \cellcolor{correct}\textbf{0.052} & \cellcolor{correct}0.096 & \cellcolor{correct}0.063 & \cellcolor{correct}0.099 & \cellcolor{correct}0.064 \\
        19 & \cellcolor{correct}\textbf{0.045} & \cellcolor{correct}0.121 & \cellcolor{correct}0.071 & \cellcolor{correct}0.132 & \cellcolor{correct}0.108 \\
        20 & \cellcolor{incorrect}\textbf{0.042} & \cellcolor{correct}0.101 & \cellcolor{incorrect}0.084 & \cellcolor{incorrect}0.073 & \cellcolor{correct}0.063 \\
        \bottomrule
    \end{tabular}
    }
    \caption{The entropy for different strategies. The minimum entropy value for each problem is in bold. Gree denotes a valid conclusion, red denotes invalid. \textit{Nan} denotes that we did not successfully parse the keyword ("Answser:") in the response.}
    \label{ent}
\end{table}

\section{Probability and Entropy Results}

The overall probabilities of Phi-4 responses for both \texttt{TruthQuest} and \texttt{ZebraLogic} are shown in Table \ref{mean&std_prob}.
Analogous entropy numbers are presented in Table \ref{mean&std_ent}.
 
\begin{table*}
 \centering
 \begin{tabular}{|l|cc|cc|cc|cc|}  % <-- Updated to match 8 columns
 \hline
 & \multicolumn{4}{|c|}{\texttt{TruthQuest}} & \multicolumn{4}{|c|}{\texttt{ZebraLogic}} \\
   
   \hline
     & \multicolumn{2}{|c|}{$P_{\text{answer}}$} 
    & \multicolumn{2}{|c|}{$P_{\text{rational}}$}  
    & \multicolumn{2}{|c|}{$P_{\text{answer}}$} 
    & \multicolumn{2}{|c|}{$P_{\text{rational}}$} \\  % Added proper column alignment
   \hline
   \textbf{Method} & Mean & Std Dev & Mean & Std Dev & Mean & Std Dev & Mean & Std Dev \\ 
   \hline
   No strategy& \textbf{0.993}& 0.038*& 0.941& 0.026*& \textbf{0.996}& 0.016*& 0.906& 0.048*\\ 
   Supposition Following   & 0.956& 0.037 & 0.943 & 0.015 & 0.992& 0.011& \textbf{0.914}& 0.021\\ 
   Chain Construction      & 0.959 & 0.037 & \textbf{0.945}& 0.011& 0.991& 0.012& 0.910& 0.022\\ 
   Compound Strategy       & 0.956 & 0.037 & 0.940 & 0.013 & 0.991& 0.011& 0.902& 0.025\\ 
   Concatenation Strategy  & 0.960 & 0.034 & 0.942 & 0.012 & 0.990& 0.013& 0.902& 0.024\\ 
   \hline
 \end{tabular}
 \caption{Mean and Standard Deviation of $P_{\text{answer}}$, $P_{\text{rational}}$  of Phi-4 model in \texttt{TruthQuest} and \texttt{ZebraLogic}. * denotes that we present twice of the calculated standard deviation, because we sample with no strategy prompt 4 times for each question.}
 \label{mean&std_prob}
\end{table*}

\begin{table*}
  \centering
  \begin{tabular}{|l|cc|cc|cc|cc|}  % <-- Updated to match 8 columns
  \hline
  & \multicolumn{4}{|c|}{\texttt{TruthQuest}} & \multicolumn{4}{|c|}{\texttt{ZebraLogic}} \\
    
    \hline
     & \multicolumn{2}{|c|}{$H_{\text{answer}}$} 
    & \multicolumn{2}{|c|}{$H_{\text{rational}}$}  
    & \multicolumn{2}{|c|}{$H_{\text{answer}}$} 
    & \multicolumn{2}{|c|}{$H_{\text{rational}}$} \\  % Added proper column alignment
    \hline
    \textbf{Method} & Mean & Std Dev & Mean & Std Dev & Mean & Std Dev & Mean & Std Dev \\ 
    \hline
    No strategy& \textbf{0.011}& 0.040*& 0.087& 0.020*& \textbf{0.006}& 0.021*& 0.142& 0.070*\\ 
    Supposition Following & 0.065 & 0.037 & 0.084 & 0.021& 0.011& 0.013& \textbf{0.130}& 0.032\\ 
    Chain Construction & 0.062 & 0.037 & \textbf{0.082}& 0.015& 0.012& 0.013& 0.136& 0.033\\ 
    Compound Strategy & 0.066 & 0.040 & 0.090 & 0.017& 0.013& 0.012& 0.150& 0.037\\ 
    Concatenation Strategy & 0.062 & 0.037 & 0.085 & 0.017& 0.015& 0.014& 0.149& 0.036\\ 
    \hline
  \end{tabular}
  \caption{Mean and Standard Deviation of $H_{\text{answer}}$, $H_{\text{rational}}$  of Phi-4 responses in \texttt{TruthQuest} and \texttt{ZebraLogic}. * denotes that we present twice of the calculated standard deviation, because we sample with no strategy prompt 4 times for each question.}
  \label{mean&std_ent}
\end{table*}

There are three conclusions we could draw from tables \ref{prob}, \ref{ent}, \ref{mean&std_prob} and \ref{mean&std_ent}:
\begin{itemize}
   \item The no-strategy prompt maintains consistently high confidence across different problems.
   \item Responses with high probability and low entropy scores are more likely to produce correct conclusions, with probability appearing to be a more reliable metric than entropy.
   \item In both \texttt{TruthQuest} and \texttt{ZebraLogic}, the no-strategy prompt exhibits a high average probability in the answer portion. In \texttt{TruthQuest}, the chain construction prompt achieves a higher average probability in the reasoning portion, whereas in \texttt{ZebraLogic}, the supposition-following prompt performs best in this regard.
\end{itemize}

\section{Combine with hyper-parameters}
A more general form of $P_{\text{combined}}$ is:

$$P_{\text{combined}} =  {P_{\text{rational}}} ^ {2(1-\lambda_p)} \times {P_{\text{answer}}} ^ {2\lambda_p}$$
In which $\lambda_p$, is a hyperparameter ranges from 0 to 1. Specially, if $\lambda_p = \frac{1}{2}$ , the formula would be the same as the main article.
Similarly, $H_{\text{combined}}$ could as be generalized to the following form:
\[
H_{\text{combined}} = (1-\lambda_e) H_{\text{rational}} + \lambda_e H_{\text{answer}}
\]
In which $\lambda_e$, is a hyperparameter ranges from 0 to 1.

\subsection{Impact of weight}
\begin{figure}[ht]
    \centering
    \includegraphics[width=0.5\textwidth]{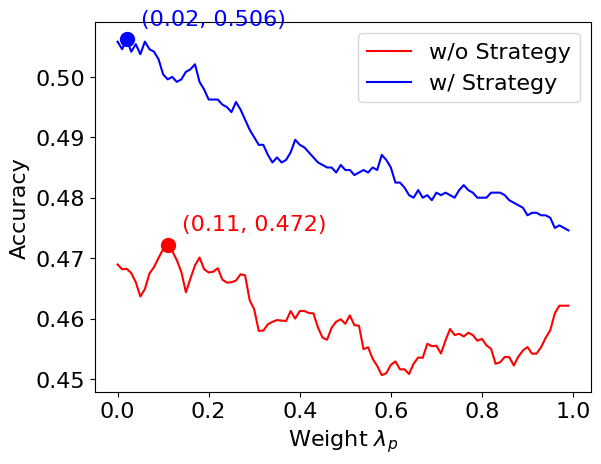}  
    \caption{The overall accuracy of Phi-4 model in TruthQuest change with maximum probability combination over different $\lambda_p$}
    \label{max_prob}
\end{figure}

\begin{figure}[ht]
    \centering
    \includegraphics[width=0.5\textwidth]{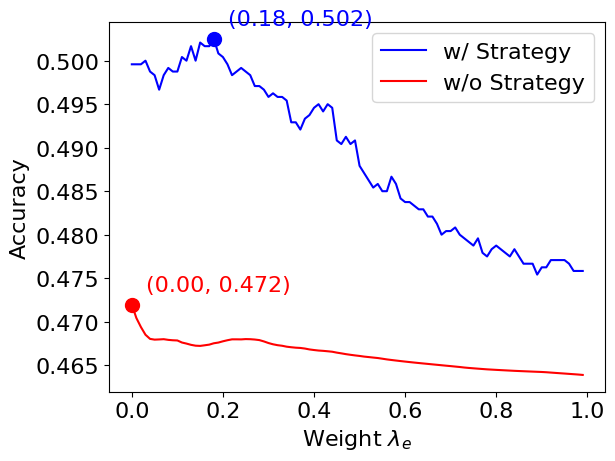}  
    \caption{The overall accuracy of Phi-4 model in TruthQuest change with minimum entropy combination over different $\lambda_e$.}
    \label{min_entropy}
\end{figure}

To assess the impact of balancing reasoning rational and answer probabilities on correctness, we compute accuracy using the \textbf{max prob@4} selection method across 100 different values of \( \lambda_p \in \{0, 0.01, 0.02, \dots\} \). The results are presented in Figure~\ref{max_prob}.  
Additionally, we evaluate how the \textbf{max entropy@4} selection varies as a function of 100 different values of \( \lambda_e \), as shown in Figure~\ref{min_entropy}. Our observations indicate a decreasing trend in \texttt{TruthQuest}, suggesting that reasoning probability or entropy plays a significant role in determining a more accurate answer.

\begin{figure}[ht]
    \centering
    \includegraphics[width=0.5\textwidth]{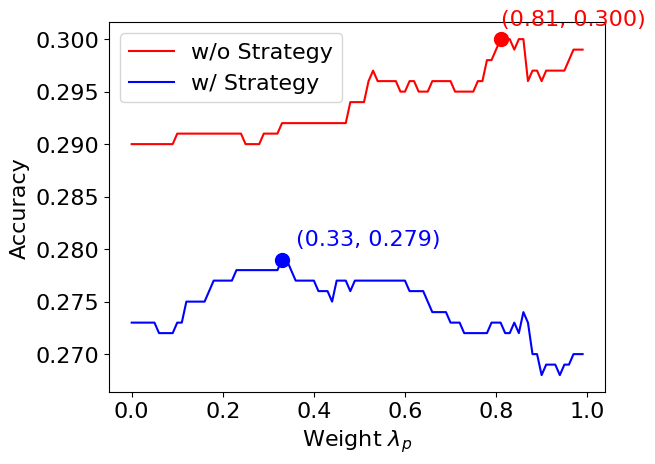}  
    \caption{The overall accuracy of Phi-4 model in ZebraLogic change with maximum probability combination over different $\lambda_p$}
    \label{max_prob_zebra}
\end{figure}

\begin{figure}[ht]
    \centering
    \includegraphics[width=0.5\textwidth]{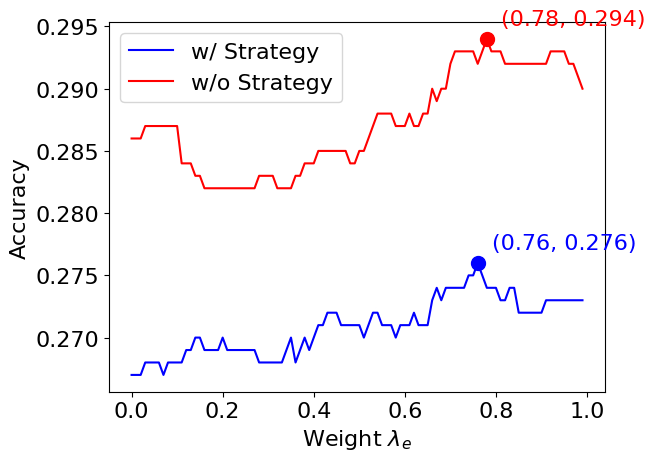}  
    \caption{The overall accuracy of Phi-4 model in ZebraLogic change with minimum entropy combination over different $\lambda_e$}
    \label{min_entropy_zebra}
\end{figure}

The plots for \texttt{ZebraLogic} are shown in Figure \ref{max_prob_zebra} and Figure \ref{min_entropy_zebra}. The tendency is totally different for this dataset. This indicates that the impact on correctness of rational and answer is specific to dataset and should be tuned accordingly.

\section{Analysis of Merging Accuracy}
\begin{figure}[ht]
    \centering
    \includegraphics[width=\linewidth]{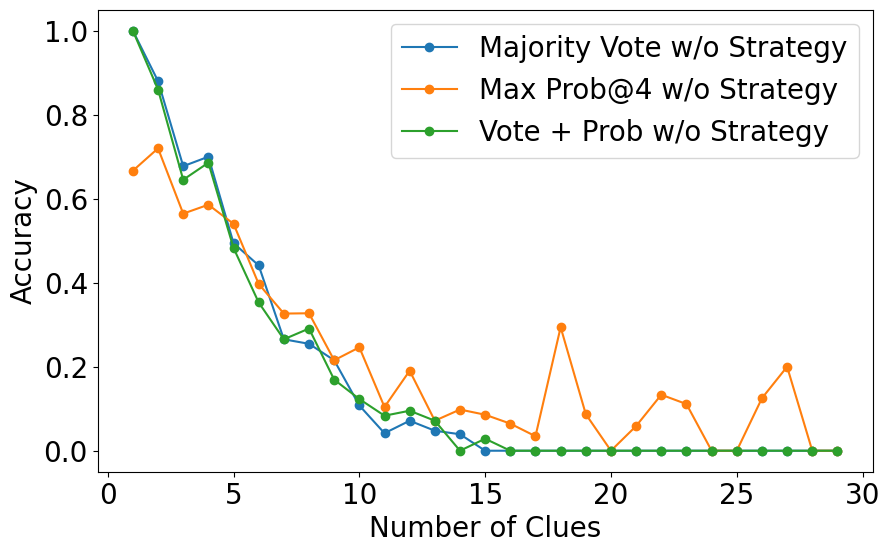}
    \caption{The accuracy distribution of different merging strategy of no strategy prompt}
    \label{fig:clue_no_strategy}
\end{figure}

\begin{figure}[ht]
    \centering
    \includegraphics[width=\linewidth]{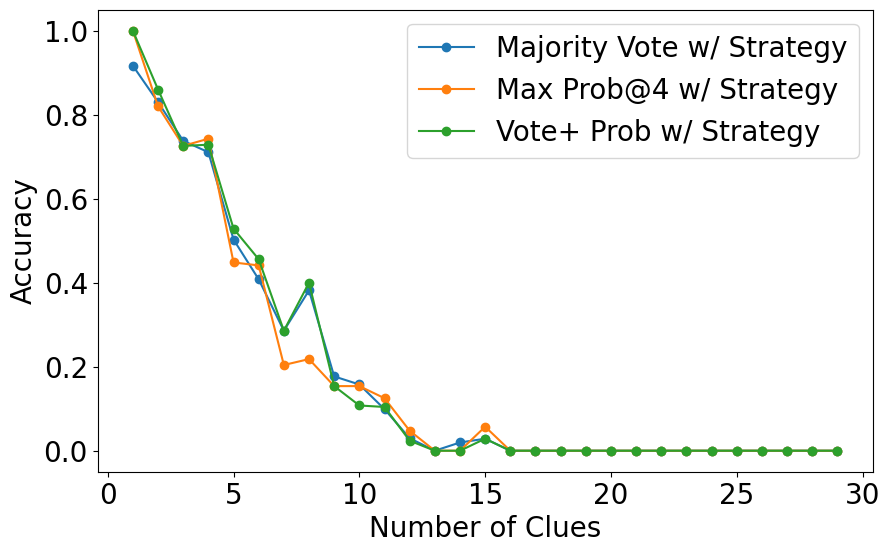}
    \caption{The accuracy distribution of different merging strategy of strategy prompt}
    \label{fig:clue_strategy}
\end{figure}

In Figure~\ref{fig:clues}, we observe that the performance of strategy prompts tends to decline as the number of clues increases. Additionally, we investigate the impact of different merging strategies on performance. The accuracy distribution of various merging strategies is presented in Figure~\ref{fig:clue_no_strategy} and Figure~\ref{fig:clue_strategy}.  

Our findings indicate that the \textbf{majority vote} approach fails to fully leverage potential valid conclusions when the number of clues is high. Conversely, the \textbf{max prob} method improves the identification of valid conclusions in cases with many clues but is more prone to selecting incorrect answers when fewer clues are available. Notably, the strategy prompt exhibits zero accuracy in scenarios with numerous clues.

\section{Impact of Number of Clues on Accuracy in \texttt{ZebraLogic}}
\begin{figure}[ht]
    \centering
    \includegraphics[width=\linewidth]{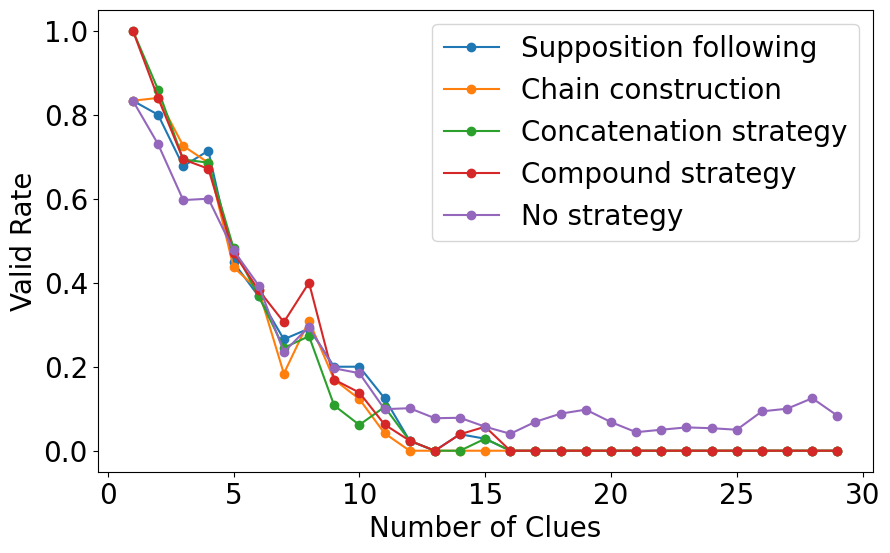}
    \caption{The accuracy distribution across different clue scale}
    \label{fig:clues}
\end{figure}

We present an analysis on why the performance of strategy prompts is lower than no strategy in \texttt{ZebraLogic}. We compute the valid rate in respective of the scales of clues, and show the result in Figure~\ref{fig:clues}. We observe that when the number of clues is lower than 8, the strategy prompt accuracy is greater or approximately similar to prompt without strategy. However, when the number of clues increases, especially larger than 16, strategy prompts perform badly. This could be a result of the limited tokens we restricted LLM to generate, and we expect this could be alleviated with the allowance of larger number of tokens. 

% In \texttt{TruthQuest}, due to the uncertainty of role of characters, the statement from each character could not be used directly. For this reason, we think \texttt{TruthQuest} is more complex than \texttt{ZebraLogic}. Therefore, we think the strategy prompt is good at handling complex logic reasoning in limited scale but become worse with increase to large scale.

\section{Prompt Details}\label{sec:prompt}
\subsection{Prompts for \texttt{TruthQuest}}
\subsubsection*{No strategy prompt} \label{zero-shot_prompt}
For the convenience of comparison, we show the prompt in previous work~\cite{mondorf-plank-2024-liar} below:
\begin{promptbox}
[INST] Your task is to solve a logical reasoning problem.\\
You are given set of statements from which you must logically deduce the identity of a set of characters.\newline
You must infer the identity of each character.\newline
First, explain your reasoning. At the end of your answer, you must clearly state the identity of each character by following the format:\newline
Answer:\\
A: ...\\
B: ...\\
C: ...\\
...\newline
\#\#\# Instruction \#\#\#\newline
Assume that there exist only two types of people: knights and knaves. Knights always tell the truth, while knaves always lie.\\
You are given the statements from \{number of characters\} characters. Based on their statements, infer who is a knight and who is a knave.\newline
End your answer by clearly stating the identity of each character in the following format:\newline
Answer:\\
A: \{knight/knave\}\\
B: \{knight/knave\}\\
C: \{knight/knave\}\newline
\#\#\# Now your turn \#\#\#\newline
Based on the following statements, infer who is a knight and who is a knave:\\
\{Question\}\newline
Let’s think step by step. [/INST]
\end{promptbox}

\subsubsection*{Supposition following prompt} \label{sup_prompt}
The prompt for supposition following for \texttt{TruthQuest} is a following:
\begin{promptbox}
[INST] Your task is to solve a logical reasoning problem.\\
You are given set of statements from which you must logically deduce the identity of a set of characters.\newline
You must infer the identity of each character. First, explain your reasoning. At the end of your answer, you must clearly state the identity of each character by following the format:\newline
Answer:\\
A: ...\\
B: ...\\
C: ...\\
...\newline
\#\#\# Instruction \#\#\#\newline
Assume that there exist only two types of people: knights and knaves. Knights always tell the truth, while knaves always lie.\\
You are given the statements from \{number of characters\} characters. Based on their statements, infer who is a knight and who is a knave.\newline
You will reason with supposition following. You employ this strategy starting with a supposition, e.g., by assuming a character is a certain type. Subsequently, trace the implications of that supposition, logically following from the premises at hand.\newline
Let’s break it down step by step:\newline
Step 1: Use numbers to define all possible propositions (proposition 1, proposition 2, ...). These propositions should be atomic, so do not include negative tones (like "not") or first-order logic (like if, then).\newline
Step 2: Make a supposition (one proposition is true or false) to test its logical implications.\newline
Step 3: Trace all the consequences of this supposition based on the premises.\newline
Step 4: If you identify any contradictions, test the alternative supposition.\newline
Step 5: If all propositions have been tested without contradiction, draw a final conclusion by following the format:\newline
Answer:\\
A: \{knight/knave\}\\
B: \{knight/knave\}\\
C: \{knight/knave\}\newline
\#\#\# Now your turn \#\#\#\newline
Based on the following statements, infer who is a knight and who is a knave:\\
\{Question\}\newline
Let’s think step by step. [/INST]
\end{promptbox}

\subsubsection*{Chain construction prompt} \label{chain_prompt}
The prompt for chain construction for \texttt{TruthQuest} is shown in main part of the paper.
% \begin{promptbox}
% [INST] Your task is to solve a logical reasoning problem.\\
% You are given a set of statements from which you must logically deduce the identity of a set of characters.\newline
% You must infer the identity of each character. First, explain your reasoning. At the end of your answer, you must clearly state the identity of each character by following the format:\newline
% Answer:\\
% A: ...\\
% B: ...\\
% C: ...\\
% ...\newline
% \#\#\# Instruction \#\#\#\newline
% Assume that there exist only two types of people: knights and knaves. Knights always tell the truth, while knaves always lie.\\
% You are given the statements from \{number of characters\} characters. Based on their statements, infer who is a knight and who is a knave.\newline
% You will reason with chain construction. You construct a chain of propositional statements derived either from the problem description or from intermediate deductions.\newline
% Let’s break it down step by step:\newline
% Step 1: Identify the logical relationships in each statement, clarifying their conditions.\newline
% Step 2: Deduce intermediate implications step by step based on the statements.\newline
% Step 3: Construct a coherent logical chain and draw a final conclusion by following the format:\newline
% Answer:\\
% A: \{knight/knave\}\\
% B: \{knight/knave\}\\
% C: \{knight/knave\}\\
% ...\newline
% \#\#\# Now your turn \#\#\#\newline
% Based on the following statements, infer who is a knight and who is a knave:\\
% \{Question\}\newline
% Let’s think step by step. [/INST]
% \end{promptbox}

\subsubsection*{Compound strategy prompt} \label{compound_prompt}
The prompt for compound strategy for \texttt{TruthQuest} is a following:
\begin{promptbox}
[INST] Your task is to solve a logical reasoning problem.\\
You are given a set of statements from which you must logically deduce the identity of a set of characters.\newline
You must infer the identity of each character. First, explain your reasoning. At the end of your answer, you must clearly state the identity of each character by following the format:\newline
Answer:\\
A: ...\\
B: ...\\
C: ...\\
...\newline
\#\#\# Instruction \#\#\#\newline
Assume that there exist only two types of people: knights and knaves. Knights always tell the truth, while knaves always lie.\\
You are given the statements from \{number of characters\} characters. Based on their statements, infer who is a knight and who is a knave.\newline
You will reason with compound strategy. You combine two or more statements to derive a new compound conclusion. This process yields a series of novel conclusions, each building upon the preceding ones. Subsequently, you trace the implications of that supposition, logically following from the premises at hand.\newline
Let’s break it down step by step:\newline
Step 1: Identify the logical relationships in each statement, clarifying their proposition.\newline
Step 2: Analyze the relationships in two statements to derive an intermediate conclusion.\newline
Step 3: Use the intermediate conclusion and another statement to build on the reasoning process.\newline
Step 4: Combine all intermediate steps logically to establish whether the conclusion follows.\newline
Step 5: Draw a final conclusion by following the format:\newline
Answer:\\
A: \{knight/knave\}\\
B: \{knight/knave\}\\
C: \{knight/knave\}\\
...\newline
\#\#\# Now your turn \#\#\#\newline
Based on the following statements, infer who is a knight and who is a knave:\\
\{Question\}\newline
Let’s think step by step. [/INST]
\end{promptbox}

\subsubsection*{Concatenation strategy prompt} \label{concat_prompt}
The prompt for concatenationd strategy for \texttt{TruthQuest} is a following:
\begin{promptbox}
[INST] Your task is to solve a logical reasoning problem.\\
You are given a set of statements from which you must logically deduce the identity of a set of characters.\newline
You must infer the identity of each character. First, explain your reasoning. At the end of your answer, you must clearly state the identity of each character by following the format:\newline
Answer:\\
A: ...\\
B: ...\\
C: ...\\
...\newline
\#\#\# Instruction \#\#\#\newline
Assume that there exist only two types of people: knights and knaves. Knights always tell the truth, while knaves always lie.\\
You are given the statements from \{number of characters\} characters. Based on their statements, infer who is a knight and who is a knave.\newline
You will reason with concatenation strategy. This entails the concatenation of two or more statements into a single conclusion encompassing the logical implications of each combined proposition.\newline
Let’s break it down step by step:\newline
Step 1: Identify initial premises suitable for concatenation. \\
Step 2: Perform concatenation by combining premises into intermediate conclusions. \\
Step 3: Repeat concatenation of intermediate conclusions as needed to build toward the final result. \\
Step 4: Draw a final conclusion.\\
Answer:\\
A: \{knight/knave\}\\
B: \{knight/knave\}\\
C: \{knight/knave\}\\
...\newline
\#\#\# Now your turn \#\#\#\newline
Based on the following statements, infer who is a knight and who is a knave:\\
\{Question\}\newline
Let’s think step by step. [/INST]
\end{promptbox}

\subsection{Prompts for \texttt{ZebraLogic}}
\subsubsection*{No strategy prompt}
The zero-shot version from the prompt of \texttt{ZebraLogic}~\cite{zebralogic2024}.
\begin{promptbox}
[INST] Your task is to solve a logical reasoning problem.\\
You are given a problem with a set of clues from which you must logically deduce the features of each house.\newline
First, explain your reasoning. At the end of your answer, you must clearly state the features of each house by following the format:\newline
Answer:\\
House 1: ...\\
House 2: ...\\
House 3: ...\\
...\newline
\#\#\# Instruction \#\#\#\newline
You are given a problem with a set of clues for \{number of houses\} houses. Each house has \{number of features\} features. You must infer the features of each house.\newline
\#\#\# Now your turn \#\#\#\newline
\{Question\}\newline
End your answer by clearly stating the features of each house in the following format:\newline
Answer:\\
\{Template\}\newline
Let’s think step by step. [/INST]
\end{promptbox}

\subsubsection*{Chain construction prompt}
We give an exemplary prompt of chain construction strategy for \texttt{ZebraLogic}. It is a direct transformation from the prompt from \texttt{TruthQuest} without changing the strategy description itselves. The other strategy prompt following the same transformation.
\begin{promptbox}
[INST] Your task is to solve a logical reasoning problem.\\
You are given a problem with a set of clues from which you must logically deduce the features of each house.\newline
First, explain your reasoning. At the end of your answer, you must clearly state the features of each house by following the format:\newline
Answer:\\
House 1: ...\\
House 2: ...\\
House 3: ...\\
...\newline
\#\#\# Instruction \#\#\#\newline
You are given a problem with a set of clues for \{number of houses\} houses. Each house has \{number of features\} features. You must infer the features of each house.\newline
You will reason with chain construction. You construct a chain of propositional statements derived either from the problem description or from intermediate deductions.\newline
Let’s break it down step by step:\newline
Step 1: Identify the logical relationships in each clue, clarifying their conditions.\newline
Step 2: Deduce intermediate implications step by step based on the clues.\newline
Step 3: Construct a coherent logical chain and draw a final conclusion.\newline
\#\#\# Now your turn \#\#\#\newline
\{Question\}\newline
End your answer by clearly stating the features of each house in the following format:\newline
Answer:\\
\{Template\}\newline
Let’s think step by step. [/INST]
\end{promptbox}
It is straightforward to adapt the transformation from \texttt{ZebraLogic} for other three strategies, so we would not show them here to save space.

\end{document}